\documentclass{article}

\usepackage[final]{corl_2026} 

\usepackage{caption}
\usepackage{xcolor} 
\usepackage{graphicx}
\usepackage{amsmath}
\usepackage{booktabs}
\usepackage{pifont}
\usepackage{wrapfig}
\usepackage{amssymb}
\usepackage{tikz}
\usepackage{pgfplots}
\pgfplotsset{compat=1.18}
\usepackage[table]{xcolor}  
\usepackage{bm}
\usepackage{algorithm}
\usepackage{algorithmic}
\usepackage{hyperref}

\title{AC-VLA: Robust Out-of-Distribution Action Execution via Compositional Learning}

\author{
  Xiaojiang Peng$^{1,\ast}$, Kai Peng$^{1,\thanks{Co-first author. $^\dagger$ Corresponding author.}}$, Jie Lu$^{1}$, Zheng Lian$^{3}$, Zitong Yu$^{4}$, Xiaobo Wang$^{2,\dagger}$\\
 $^1$Shenzhen Technology University, \\$^2$Shenzhen University of Advanced Technology,\\ $^3$Tongji University, ~$^4$Great Bay University
}

\begin{document}
\maketitle


%
\begin{abstract}
Vision-Language-Action (VLA) models excel at end-to-end robotic manipulation but struggle with out-of-distribution (OOD) generalization when familiar sub-tasks are recombined in unseen configurations. We identify two mutually reinforcing failure modes: \emph{trajectory overfitting}, where models overfit to holistic trajectory patterns rather than compositional sub-skill semantics; and \emph{perceptual shortcut}, where action tokens over-rely on wrist-view textures at the expense of global spatial grounding. To address both, we introduce \textbf{AC-VLA}, a plug-and-play Action Compositional learning framework comprising two architecture-agnostic components: \textbf{(i)} a compositional learning module that uses an LLM-driven instruction decomposer and a proprioceptive trajectory aligner to generate dense sub-task supervision, followed by mixed training on complete demonstrations and decomposed data to endow the model with compositional generalization; and \textbf{(ii)} a state-conditioned asymmetric masking strategy that suppresses wrist-view inputs during closed-gripper phases, enforcing global semantic grounding. All components are architectural modification-free and directly integrable into any VLA backbone. Instantiated on $\pi_{0.5}$ and evaluated on LIBERO and LIBERO-OOD benchmarks, AC-VLA achieves a $\sim$28\% absolute improvement on compositional OOD tasks while maintaining near-perfect in-distribution performance. The source code will be made publicly available on \href{https://ac-vla.github.io/}{this page}.
\end{abstract}




\section{Introduction}
Vision-Language-Action (VLA) models~\cite{li2025cogvla,bjorck2025gr00t,physicalintelligence2024pi0,zitkovich2023rt2,kim2024openvla,wen2024tinyvla,shukor2025smolvla,brohan2023rt1,ghosh2024octo,liu2024rdt,liang2026adaptive} have achieved widespread adoption in robotic manipulation, owing to their ability to map language and vision directly to continuous control actions through large-scale pre-training. However, these models exhibit marked degradation when faced with novel instructions that recombine familiar sub-tasks in unseen configurations, even within the training distribution~\cite{fang2026when}.

This failure mainly stems from reliance on spurious correlations between holistic task descriptions and specific trajectory patterns, rather than a genuine compositional understanding of sub-skills. 
Simply scaling up training data is unlikely to close this gap efficiently~\cite{lindata}, as the core deficit lies in the model's failure to decompose and recompose learned primitives.  Hierarchical paradigms~\cite{robobrain2025,ahn2022saycan,huang2022inner,driess2023palme} decouple planning from execution to mitigate this issue, yet the resulting modularity prevents high-level reasoning from adapting to real-time sensory feedback, disrupting fluid control. Stage-wise VLA methods~\cite{longvla2025,atomskill2024} similarly introduce rigid sub-task boundaries that cause error accumulation and abrupt transitions.

We characterize this as a \emph{compositional generalization} challenge: the agent's ability to execute novel task instructions by recombining familiar sub-skills (e.g., performing a `pick' and `place' sequence in a new spatial configuration or with unseen object-verb pairings). 
When facing out-of-distribution tasks, we identify that existing VLA models may suffer from two mutually reinforcing failure modes: \textit{trajectory overfitting} and \textit{perception shortcuts}.

\begin{wrapfigure}{r}{0.6\linewidth}
    \centering
    \vspace{-1.5em}
    \includegraphics[width=1\linewidth]{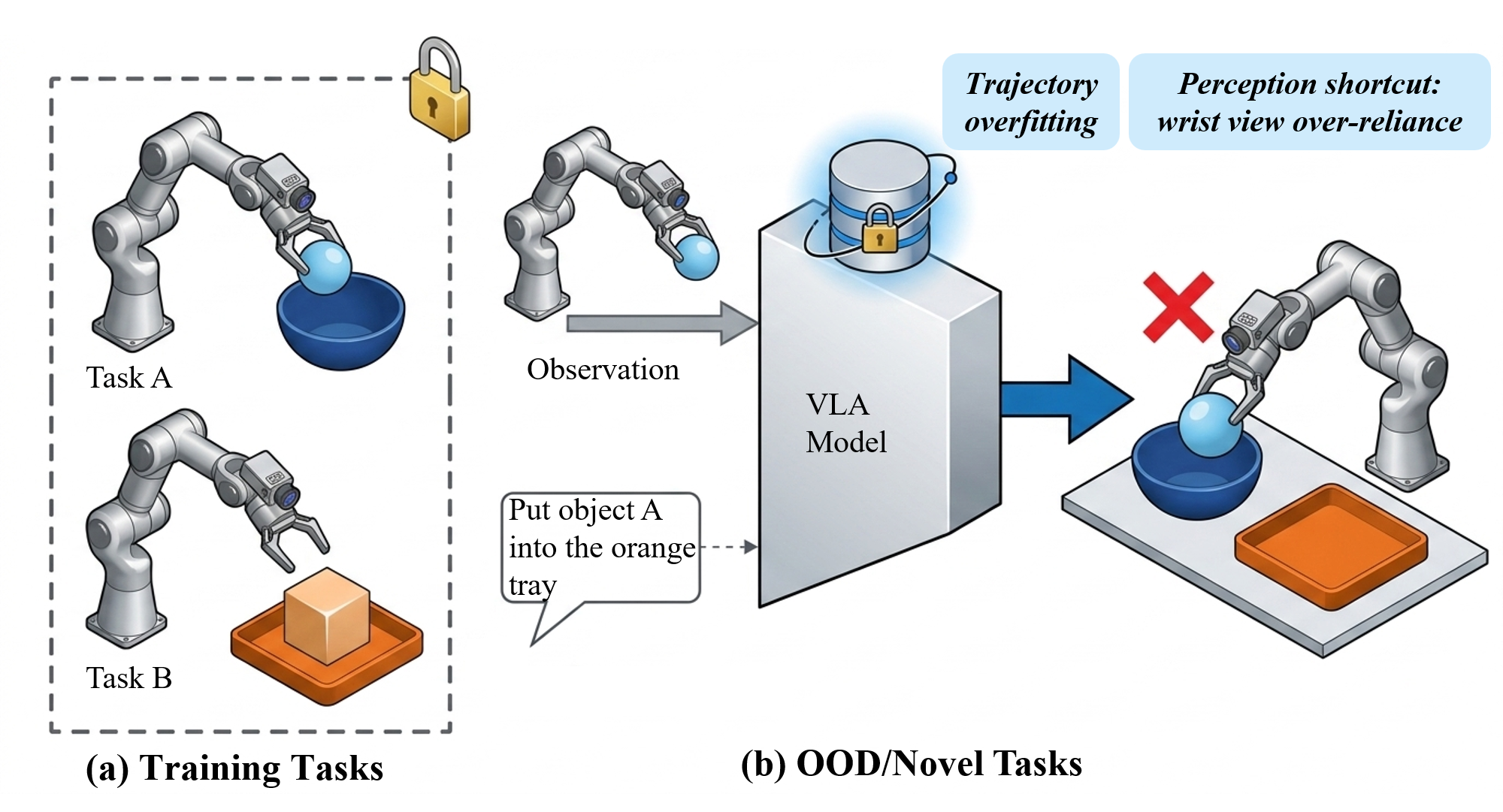}
    \caption{Out-of-distribution compositional actions confuse vision and language modalities. (a) Training tasks: object A$\rightarrow$target A and object B$\rightarrow$target B. (b) Novel task: object A$\rightarrow$target B. Since all training examples map ``grasp object A'' to target A, the model faces a token-level conflict between visual cues (object A) and language instructions (target B), leading to action prediction confusion.}
    \label{fig:moti}
    \vspace{-0.5em}
\end{wrapfigure}

As illustrated in Figure \ref{fig:moti}, trajectory overfitting manifests when models memorize complete motion sequences associated with seen task combinations, failing to disentangle reusable sub-skills such as “grasp A” and “place at B.” When confronted with a recombined instruction, the model attempts to replicate a familiar holistic trajectory rather than dynamically composing the relevant primitives, leading to rigid and incorrect behaviors. In parallel, perceptual shortcut arises from an excessive dependence on local wrist-view textures, which provide strong but spurious visual cues tightly coupled to specific objects during grasping. When the task configuration changes—e.g., the same object must now be placed at a different target—the model remains anchored to these egocentric features and neglects the global spatial layout required for accurate placement. This dual breakdown, semantic overfitting at the trajectory level and visual shortcut at the perception level, severely undermines compositional generalization in current VLA models.

To facilitate compositional generalization in VLA models, we introduce AC-VLA, a plug-and-play Action Compositional learning framework that equips VLA models with the ability to recombine sub-skills for robust out-of-distribution (OOD) execution. AC-VLA consists of two architecture-agnostic components. First, a \textit{compositional learning module} decomposes demonstrations into reusable sub-task units: an LLM-driven instruction decomposer segments complex instructions into granular sub-task descriptions, while a proprioceptive trajectory aligner partitions the corresponding trajectory segments, generating dense sub-task supervision. A mixed training strategy then jointly trains on complete task demonstrations and the decomposed sub-task data, systematically preserving holistic task coherence while instilling compositional generalization. Second, a \textit{state-conditioned asymmetric masking} strategy suppresses wrist-view inputs during closed-gripper phases, forcing the model to develop robust global spatial grounding instead of over-relying on local wrist-view textures, thereby substantially improving generalization to OOD scenarios.


\section{Related Work}
\label{sec:Related Work}

\noindent\textbf{Vision-Language-Action Models.}~~Vision-Language-Action (VLA) models integrate visual perception, language understanding, and action generation to enable end-to-end robotic control~\cite{li2025cogvla,bjorck2025gr00t,physicalintelligence2024pi0,zitkovich2023rt2,kim2024openvla,wen2024tinyvla,shukor2025smolvla,brohan2023rt1,ghosh2024octo,liu2024rdt,li2024roboflamingo,qu2025spatialvla,doshi2024crossformer}. Leveraging internet-scale pre-training on vision-language models~\cite{bai2023qwenvl,radford2021clip,liu2024llava,alayrac2022flamingo} and diverse large-scale robot datasets~\cite{openx2024,khazatsky2024droid,liu2023libero}, these models exhibit versatile instruction-following and cross-embodiment generalization. Despite these advances, a core limitation of current VLA paradigms is that they supervise complete task demonstrations as holistic trajectories paired with a single, coarse language description. This coarse alignment encourages memorization of in-distribution trajectory patterns, significantly degrading performance on out-of-distribution tasks~\cite{fang2026when,lindata}. Our work directly addresses this limitation through structured compositional learning, without any architectural modification.

\noindent\textbf{Hierarchical Task Segmentation.}~~Hierarchical task segmentation aims to decompose complex, long-horizon manipulation demonstrations into structured, reusable primitives that support compositional generalization. Early approaches employ high-level planners to break down natural language instructions into sub-goal sequences, either via LLM-based language decomposition~\cite{ahn2022saycan,huang2022inner,liang2023code} or visual subgoal generation~\cite{black2024zero,chen2025goal,kang2025incorporating}. These modular designs introduce a structural decoupling between linguistic planning and low-level action execution, hindering joint optimization within unified VLA frameworks. More recent methods embed hierarchical segmentation directly into end-to-end VLA training~\cite{longvla2025,atomskill2024}, providing dense supervision and fine-grained alignment between language instructions and visuo-motor segments. However, existing approaches typically rely on fixed-length temporal partitioning, which compromises semantic fidelity, or require task-specific engineering that limits scalability. Our work addresses this gap with \textit{an automated, proprioceptive-driven aligner that synchronizes LLM-decomposed semantic instructions with robot proprioceptive states—leveraging gripper-state transitions and cumulative displacement} as natural semantic boundaries—yielding high-density supervision without manual annotation or task-specific design. In contrast to LLM-based planners~\cite{driess2023palme,singh2023progprompt,wang2023voyager} that depend on online inference or explicit skill libraries~\cite{garg2022lisa,chen2021dt,zhao2025cot}, our decomposition operates fully offline and integrates seamlessly into any VLA training pipeline.

\noindent\textbf{Out-of-Distribution Generalization.}~~Enabling agents to recombine acquired motor primitives for executing novel task instructions beyond the training distribution is a central challenge in manipulation. Recent works explore this from complementary angles: sub-skill-oriented training improves fine-grained action coverage~\cite{atomskill2024}, stage-wise sequential modeling structures task execution into discrete learnable phases~\cite{longvla2025}, and counterfactual analysis reveals the visual-shortcut failure modes that impede language grounding~\cite{fang2026when}. Nevertheless, training on decomposed or stage-partitioned data alone overextends model capacity, dilutes holistic task knowledge, and causes significant in-distribution performance degradation. 
Concurrent works explore data augmentation~\cite{wang2023mimicplay,bharadhwaj2024roboagent} and strong policy learning baselines including ACT~\cite{zhao2023act}, UMI~\cite{chi2024umi}, Diffusion Policy~\cite{chi2023diffusionpolicy}, 3D Diffusion Policy~\cite{ze2024dp3}, and human-in-the-loop RL~\cite{luo2024hilserl}, which offer orthogonal improvements to our supervision restructuring approach.
\textit{Our mixed training paradigm explicitly addresses this stability-plasticity dilemma, and our state-conditioned asymmetric masking provides a complementary mechanism to mitigate perceptual shortcuts. }

\section{Method}

\subsection{Overview}
\begin{figure}[t]
        \centering
        \includegraphics[width=1\linewidth]{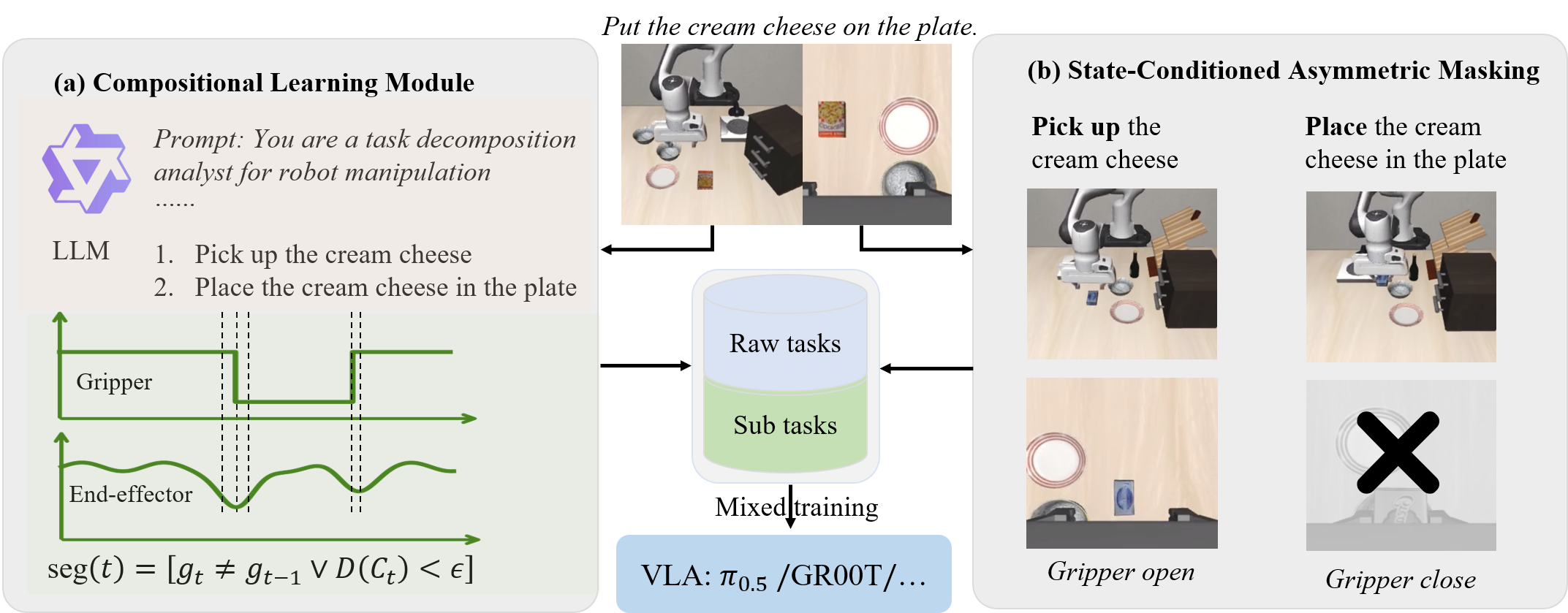}
        \caption{Overview of the AC-VLA framework. (a) Compositional learning module: full demonstrations are decomposed into sub-tasks via LLM-based instruction parsing and proprioceptive trajectory alignment; the model is then trained with a mixture of complete and decomposed data. (b) Asymmetric masking module: wrist-camera views are suppressed during closed-gripper phases at training time, enforcing global spatial grounding.}
        \label{fig:pipeline}
        \vspace{-1em}
    \end{figure}

We present the AC-VLA framework, as illustrated in Figure~\ref{fig:pipeline}. Given a raw dataset of manipulation demonstrations, our framework equips any VLA backbone with two plug-and-play components that operate without architectural modification. First, a \textbf{compositional learning module} (Section~\ref{sec:comp}) automatically decomposes each demonstration into semantically meaningful sub-task units via LLM-based instruction parsing and proprioceptive trajectory alignment, and then applies a mixed training strategy on the original and decomposed data to instill compositional generalization. Second, a \textbf{state-conditioned asymmetric masking} strategy (Section~\ref{sec:mask}) blinds wrist-camera views during closed-gripper phases for training, breaking spurious visual shortcuts and enforcing global spatial grounding. Together, these components enable robust out-of-distribution action execution in the inference phase while preserving in-distribution performance.

\subsection{Compositional Learning}
\label{sec:comp}
To address trajectory overfitting, we design a compositional learning module that decouples holistic demonstrations into reusable primitives and learns from both granularities. This module consists of two stages: collaborative decomposition and mixed training.

\subsubsection{Collaborative Decomposition with Language and Proprioceptive Trajectory}
We design a collaborative decomposition scheme to synchronize sub-task descriptions and proprioceptive actions, as illustrated in Figure \ref{fig:pipeline}(a).
Given a demonstration with a language instruction $\ell$ and a trajectory $\xi = \{(\mathbf{o}_t, \mathbf{a}_t)\}_{t=1}^{T}$, where $\mathbf{o}_t$ includes third-view and wrist-view images, and $\mathbf{a}_t$ encompasses actions such as end-effector pose and gripper status, we formulate it as a set of sub-task segments as:
\begin{equation}
    \xi = \{s_i\}_{i=1}^{N}, \quad s_i = (\ell_n, \xi_n).
\end{equation}

An LLM first decomposes $\ell$ into an ordered sequence of sub-task descriptions $\{\ell_1, \ell_2, ..., \ell_N\}$, and $N$ is dynamically determined by the LLM based on the semantic complexity of $\ell$.

To synchronize these textual sub-tasks with physical trajectory, we introduce a \textit{proprioceptive-driven aligner} that partitions $\xi$ using naturally occurring physical cues: gripper-state transitions (open $\rightarrow$ close, close $\rightarrow$ open) and chunk-wise average end-effector displacement.  Let $\mathbf{a}_t = (\mathbf{p}_t, g_t)$ denote the action at timestep $t$, where $\mathbf{p}_t \in \mathbb{R}^6$ is the end-effector pose and $g_t \in \{0,1\}$ is the gripper state. For an action chunk $A_t = \{\mathbf{a}_t, \mathbf{a}_{t+1}, \ldots, \mathbf{a}_{t+H-1}\}$ of length $H$, the chunk-wise average end-effector displacement and sub-task boundary condition are defined as:
\begin{equation}
\label{eq:seg}
    D(A_t) = \frac{1}{H}\sum_{i=1}^{H-1} \|\mathbf{p}_{t+i} - \mathbf{p}_{t+i-1}\|_2, \quad
    \text{seg}(t) = \bigl[g_t \neq g_{t-1} \lor D(A_t) < \epsilon\bigr].
\end{equation}
where $\text{seg}(t) = \texttt{True}$ indicates that timestep $t$ is identified as a sub-task boundary, and $\epsilon$ is a displacement threshold determined via clustering on the training trajectories. A gripper state transition ($g_t \neq g_{t-1}$) naturally marks the transition between grasping and transport phases, while a low displacement identifies approach-to-contact or placement events. \textit{Notably, we segment the full trajectory in a chunk-wise manner; thus, a “pick” or “place” sub-task can be obtained whenever at least one segmentation point exists in a chunk.}
The aligned pairs $(\ell_k, \xi_k)$ form a dense sub-task supervision dataset $\mathcal{D}_{\text{sub}}$, constructed fully offline without manual annotation. This preservation of linguistic granularity retains the rich compositional diversity of natural language and prevents the semantic sparsity common in stage-wise models that rely on coarse phase labels.

\subsubsection{Mixed Training: An Instance of Diffusion-based Policy}
Given the full-demonstration dataset $\mathcal{D}_{\text{full}}$ and the decomposed sub-task dataset $\mathcal{D}_{\text{sub}}$, we construct a mixed dataset \(\mathcal{D}_{\text{mix}} = \mathcal{D}_{\text{full}} \cup \mathcal{D}_{\text{sub}}\) by merging complete demonstrations with decomposed sub-task segments. At each training step, mini-batches are sampled from \(\mathcal{D}_{\text{full}}\) and \(\mathcal{D}_{\text{sub}}\) with a fixed ratio. Consequently, learning from full trajectories preserves coherent long-horizon execution, while sub-task supervision explicitly trains the model to recombine primitives, resolving the stability-plasticity dilemma. Without loss of generality, 
we instantiate this mixed training paradigm on a flow-matching VLA backbone \(\pi_{0.5}\)~\cite{hejna2025pi05}, which consists of a Vision-Language Model (VLM) that produces vision-language features \(\bm{\phi}_t\) as conditioning input, and a Diffusion Transformer (DiT) action head \(\mathbf{V}_\theta\) for flow matching. Following action chunking~\cite{zhao2023act,chi2024umi,chi2023diffusionpolicy,ze2024dp3}, the model predicts a sequence of \(H\) future actions $A'_t \in \mathbb{R}^{d \times H}$ to improve temporal coherence. Given a flow-matching timestep \(\tau \in [0,1]\) and noise \(\bm{\epsilon} \sim \mathcal{N}(0, \mathbf{I})\), the noised action chunk is \(A_t^{(\tau)} = \tau A_t + (1-\tau) \bm{\epsilon}\). Together with a robot state embedding \(\bm{q}_t\), the action head minimizes the flow-matching objective:
\begin{equation}
\mathcal{L}(\theta) = \mathbb{E}_\tau \left[ \big\| \mathbf{V}_\theta(\bm{\phi}_t, A_t^{(\tau)}, \bm{q}_t) - (A_t - \bm{\epsilon}) \big\|^2 \right].
\end{equation}

At inference, an action chunk is generated from random noise \(A_t^{(0)} \sim \mathcal{N}(0, \mathbf{I})\) via \(K\)-step forward Euler integration:
\begin{equation}
A_t^{(\tau + 1/K)} = A_t^{(\tau)} + \frac{1}{K} \mathbf{V}_\theta(\bm{\phi}_t, A_t^{(\tau)}, \bm{q}_t).
\end{equation}

\subsection{State-Conditioned Asymmetric Masking} 
\label{sec:mask}
\textbf{Evidence of perception shortcuts}.
We first demonstrate the existence of perceptual shortcuts through both qualitative visualization and quantitative performance comparisons on the raw LIBERO dataset with the popular \(\pi_{0.5}\)~\cite{hejna2025pi05} VLA model. 

\begin{wrapfigure}{r}{0.6\linewidth}
\vspace{-1.5em}
    \centering
    \captionof{table}{The evidence of perception shortcuts from an ablation study on camera views on LIBERO.}
    \begin{tabular}{lccccc}
        \toprule
        \textbf{Views} & \textbf{Spatial} & \textbf{Goal} & \textbf{Object} & \textbf{Long} \\
        \midrule
        w/ both                & 98.8\% & 98.5\%  & 99.3\% & 92.9\% \\
        w/o third-person           & 52.2\% & 77.7\%  & 68.9\% & 22.2\% \\
        \rowcolor{gray!20}  w/o wrist        & 1.9\% & 1.5\%  & 2.9\%  & 0.1\% \\
        \bottomrule
    \end{tabular}     
    \label{tab:view_ablation}
    
    \centering
    \vspace{-0.5em}
    \includegraphics[width=1\linewidth]{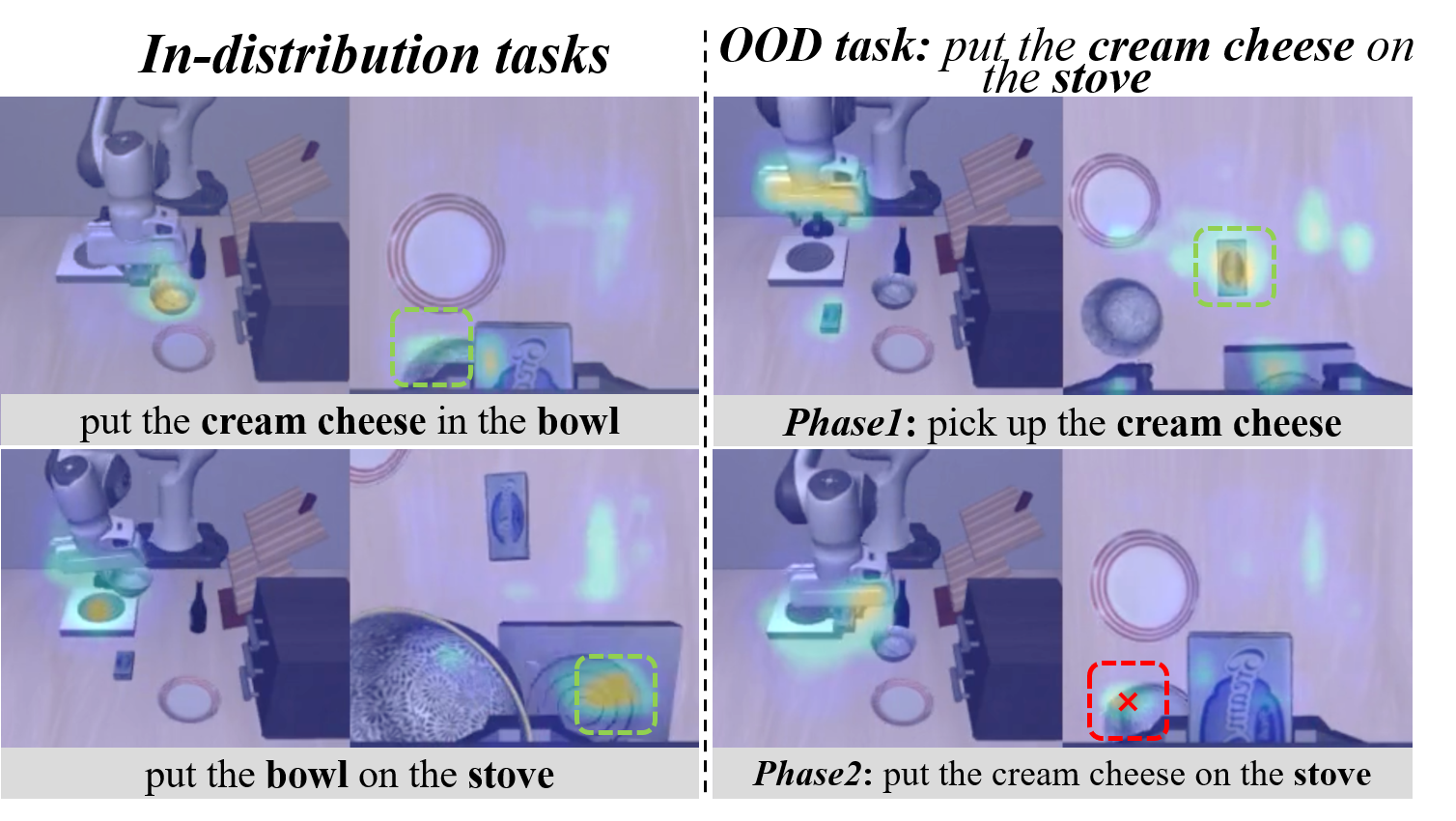}
    \captionof{figure}{Evidence of perception shortcuts in model attention maps.}
    \label{fig:maskMoti}
    \vspace{-0.5em}
\end{wrapfigure}

As shown in Table~\ref{tab:view_ablation}, removing the wrist camera causes catastrophic performance collapse across all task categories (Goal, Object, Spatial, Long), with success rates dropping to near zero. Meanwhile, removing the third-person view also substantially degrades performance, albeit to a lesser extent, revealing that the model heavily relies on both views but uses wrist-level textures as an indispensable shortcut.

To further investigate the model's behavior under OOD conditions, we visualize token-level attention scores extracted from the DiT action head on both view images. Specifically, we average the attention maps across all layers of the DiT to obtain a unified attention map per view, as shown in Figure~\ref{fig:maskMoti}. For in-distribution tasks, the attention in the ``place'' phase correctly concentrates on the intended target region. In contrast, for the OOD task, while the ``pick'' phase attention remains semantically appropriate, the ``place'' phase still exhibits strong attention to the original target (e.g., the bowl) instead of the newly instructed destination. This indicates a persistent perceptual shortcut inherited from training spurious correlations.

\textbf{Asymmetric Masking}.
The above findings motivate our state‑conditioned asymmetric masking strategy, which deliberately suppresses wrist‑view inputs to force the model to rely on global spatial context and correctly follow language directives. Specifically, we leverage the Collaborative Decomposition module to identify all “place” phase segments and mask the wrist-view tokens in the attention computation (\textit{i.e.}, mask attention) during training.
We observe that this compels the model to infer spatial relationships from third‑view observations and language instructions, rather than tracking object textures in the egocentric view. Critically, wrist inputs remain intact when the gripper is open (e.g., during approaching and grasping phases), preserving the fine‑grained feedback essential for precise manipulation. Conditioned on the proprioceptive gripper state, this asymmetric design severs spurious correlations between wrist textures and task configurations, encouraging the model to learn a globally grounded action policy. Masking is applied online during training and requires no architectural modifications or changes to the inference procedure.

\section{Experiments}
In this section, we design simulation and real-world experiments to answer the following research questions:
\textbf{RQ1}: How does our proposed paradigm enhance state-of-the-art (SOTA) models on OOD tasks?
\textbf{RQ2}: How does the composition learning module affect performance on both in-distribution and OOD tasks?
\textbf{RQ3}: How does asymmetric masking contribute to our method? 

\subsection{Experiment Setup}
\label{Experiment Setup}
\noindent\textbf{Simulation and real-world experiments}.
We evaluate our method on the LIBERO benchmark~\cite{liu2023libero}, a standardized suite for lifelong robot learning that provides diverse tabletop manipulation tasks across four categories: LIBERO-Spatial (spatial relationship), LIBERO-Goal (goal-conditioned), LIBERO-Object (object-level), and LIBERO-Long (long-horizon chaining). To assess out-of-distribution (OOD) generalization, we adopt the LIBERO-OOD benchmark~\cite{ood2026}, which consists of the LIBERO-Spatial-OOD and LIBERO-Goal-OOD suites, each containing 20 novel compositional tasks. 
In addition, we design four real-world tasks and evaluate AC-VLA on two OOD variants of these tasks (Sec.~\ref{sec:realworld}). 

For language instruction decomposition, we employ the Qwen-3.5-Flash language model to decompose the original holistic task descriptions into granular sub-task text sequences. We primarily instantiate our method on the GR00T and \(\pi_{0.5}\) backbones, using their default hyperparameters. All training experiments are run on 4 NVIDIA A100 80GB GPUs, and inference is performed on an NVIDIA RTX 5880 Ada 40GB. For more details, please refer to the appendix materials.

\noindent\textbf{Evaluation protocol}.
Each task is evaluated over 3 random seeds, each with 50 episodes (150 episodes total per task). We report the mean success rate (\%) over all tasks within each suite. 

    \begin{table}[t]
    \centering
    \caption{Performance comparison on LIBERO benchmark, including both in-distribution and out-of-distribution suites. Results are success rates (\%) averaged across tasks.}
    \label{tab:libero_results}
    \small
    \begin{tabular}{lcccc|cc|c}
        \toprule
        \textbf{Model} & \textbf{Spatial} & \textbf{Goal} & \textbf{Object} & \textbf{Long} & \textbf{Spatial OOD} & \textbf{Goal OOD} &  \textbf{AVG} \\
        \midrule
       UniVLA~\cite{univla2025}                   & 96.5   & 95.6   & 96.8   & 92.0   & 11.0   & 32.0   & 70.7  \\
        OpenVLA-OFT~\cite{kim2024openvlaoft}                        & 97.6   & 97.9   & 98.4   & 94.5   & 0.0   & 1.0   & 64.9  \\
         \rowcolor{pink!20} $\pi_{0.5}$~\cite{hejna2025pi05}                & 98.8  & 98.5   & \textbf{99.3 } & 92.9   & 35.5   & 46.6   & 78.6  \\
         \rowcolor{pink!20}  Spatial Forcing-$\pi_{0.5}$~\cite{li2025spatial} & \textbf{99.4 } & \textbf{99.6 } & 98.8   & \textbf{96.0 } & 48.3   & 57.8  & 83.3  \\
         \rowcolor{gray!20}  GR00T-N1~\cite{bjorck2025gr00t}                      & 94.1   & 98.3   & 99.1   & 93.2   & 17.9   & 24.1   & 71.1  \\
         \rowcolor{pink!20}  \textbf{AC-VLA} ($\pi_{0.5}$)        & 98.0   & 97.7   & 98.4   & 92.4   & \textbf{64.2}(\textcolor{red}{+28.7}) & \textbf{73.3}(\textcolor{red}{+26.7})  & \textbf{87.3}(\textcolor{red}{+8.7})\\
       \rowcolor{gray!20}  \textbf{AC-VLA} (GR00T-N1)        & 95.4   & 96.7   & 99.1   & 92.3   & 36.4(\textcolor{red}{+18.5})  & 44.0(\textcolor{red}{+19.9})    &  77.3(\textcolor{red}{+6.2})  \\ 
        \bottomrule
    \end{tabular}
    \vspace{-1.5em}
    \end{table}
    
\subsection{Comparison with State-of-the-Art Policies}
We instantiate AC-VLA on both the \(\pi_{0.5}\)~\cite{hejna2025pi05} and GR00T-N1~\cite{bjorck2025gr00t} backbones. We compare against the following baselines: \textbf{SF-\(\pi_{0.5}\)} (Spatial Forcing)~\cite{li2025spatial}, an architecture-agnostic framework that enhances \(\pi_{0.5}\) with spatial visual alignment without structural modifications; \textbf{GR00T-N1.7}, a generalist humanoid VLA model; \textbf{UniVLA}~\cite{univla2025}, a universal VLA with a task-centric latent action design; and \textbf{OpenVLA-OFT}~\cite{kim2024openvlaoft}, an optimized fine-tuning variant of OpenVLA~\cite{kim2024openvla}.

As shown in Table~\ref{tab:libero_results}, all baseline models achieve near-perfect success rates on the standard LIBERO in-distribution suites, yet suffer severe OOD degradation. OpenVLA-OFT, for instance, drops to 0.0 and 1.0 on the Spatial-OOD and Goal-OOD suites, respectively. SF-\(\pi_{0.5}\), which forces VLA middle layers to align extra geometric representations, attains moderate OOD improvements over the vanilla \(\pi_{0.5}\), suggesting that explicit spatial vision alignment can partially mitigate perceptual shortcuts. Specifically, SF-\(\pi_{0.5}\) improves \(\pi_{0.5}\) on the two OOD suites by \textbf{12.8/11.2}, whereas our method yields gains of \textbf{28.7/26.7}. With the generalist
humanoid robot model GR00T-N1, our method also obtains around 20 absolute gains on OOD tasks. We note that 
 \(\pi\)‑TLI~\cite{ood2026}  relies on combining two inference in-distribution tasks and reports 85/81 success rates on the OOD suites, but it involves manually‑tuned hyperparameters (e.g., interpolation speed) and requires two rounds of inference.  Overall, our AC-VLA establishes new state-of-the-art OOD results (64.2/73.3 on Spatial/Goal OOD) while preserving strong in-distribution competence, achieving the best overall average of 87.3. In the following, unless otherwise specified, we use AC‑VLA to denote AC-VLA(\(\pi_{0.5}\)).
 
\subsection{Ablation Analyses}
\label{sec:ablation}
To answer \textbf{RQ2} and \textbf{RQ3}, we conduct a quantitative ablation study on both in-distribution (In-D) and out-of-distribution (OOD) tasks, with results summarized in Table~\ref{tab:ablation}.

\textbf{Compositional learning}. 
The first three rows present the effect of our compositional learning module. Training exclusively on raw demonstrations (row 1) yields strong in-distribution performance (97.4) but severe OOD degradation (35.5/46.6). Conversely, training solely on decomposed sub-task data (row 2) improves OOD to 54.8/68.6 at the cost of catastrophic in-distribution collapse (61.4), as the model loses long-horizon coherence. Our mixed training paradigm (row 3) combines raw and sub-task supervision, successfully resolving this stability-plasticity dilemma.

\begin{wraptable}{r}{0.55\linewidth}
\vspace{-1.0em}
    \centering
    \caption{Ablation study on both In-D and OOD tasks. For In-D performance, we report the mean success rate (\%) over four suites.}  
    \resizebox{\linewidth}{!}{
    \begin{tabular}{ccc|ccc}
        \toprule
        \textbf{Raw-task} & \textbf{Sub-task} & \textbf{Mask} & \textbf{In-D} & \textbf{Spatial OOD} & \textbf{Goal OOD} \\
        \midrule
        \rowcolor{pink!20}\ding{51} &           &           & \textbf{97.4}  & 35.5 & 46.6 \\
         \rowcolor{pink!20}  & \ding{51} &           &61.4 & 54.8 & 68.6 \\
        
        \rowcolor{pink!20}\ding{51} & \ding{51} &           & 96.6 & 51.6 & 67.5 \\
        \rowcolor{gray!20}\ding{51} &           & \ding{51} & 96.5 & 47.3 & 67.0 \\
         \rowcolor{gray!20}\ding{51} & \ding{51} & \ding{51} & 96.7 & \textbf{64.2} & \textbf{73.3} \\
        \bottomrule
    \end{tabular}}
     \label{tab:ablation}
     \vspace{-1.0em}
\end{wraptable}

\textbf{Asymmetric masking}.
The last three rows evaluate the contribution of state-conditioned asymmetric masking. Adding masking to raw-demonstration training (row 4) improves OOD performance from 35.5/46.6 to 47.3/67.0, with a negligible drop in In-D performance. This confirms that suppressing wrist-view inputs during ``place'' phases effectively mitigates perceptual shortcuts and forces global spatial grounding. When combined with compositional learning (row 5), the two components boost each other: OOD success rates rise further to 64.2 and 73.3, while in-distribution performance remains high at 96.7. These results indicate that trajectory overfitting and perceptual shortcuts are complementary failure modes, and addressing both is necessary for robust compositional generalization.

\begin{wrapfigure}{r}{0.6\linewidth}
    \centering
    \vspace{-1.5em}
    \includegraphics[width=1\linewidth]{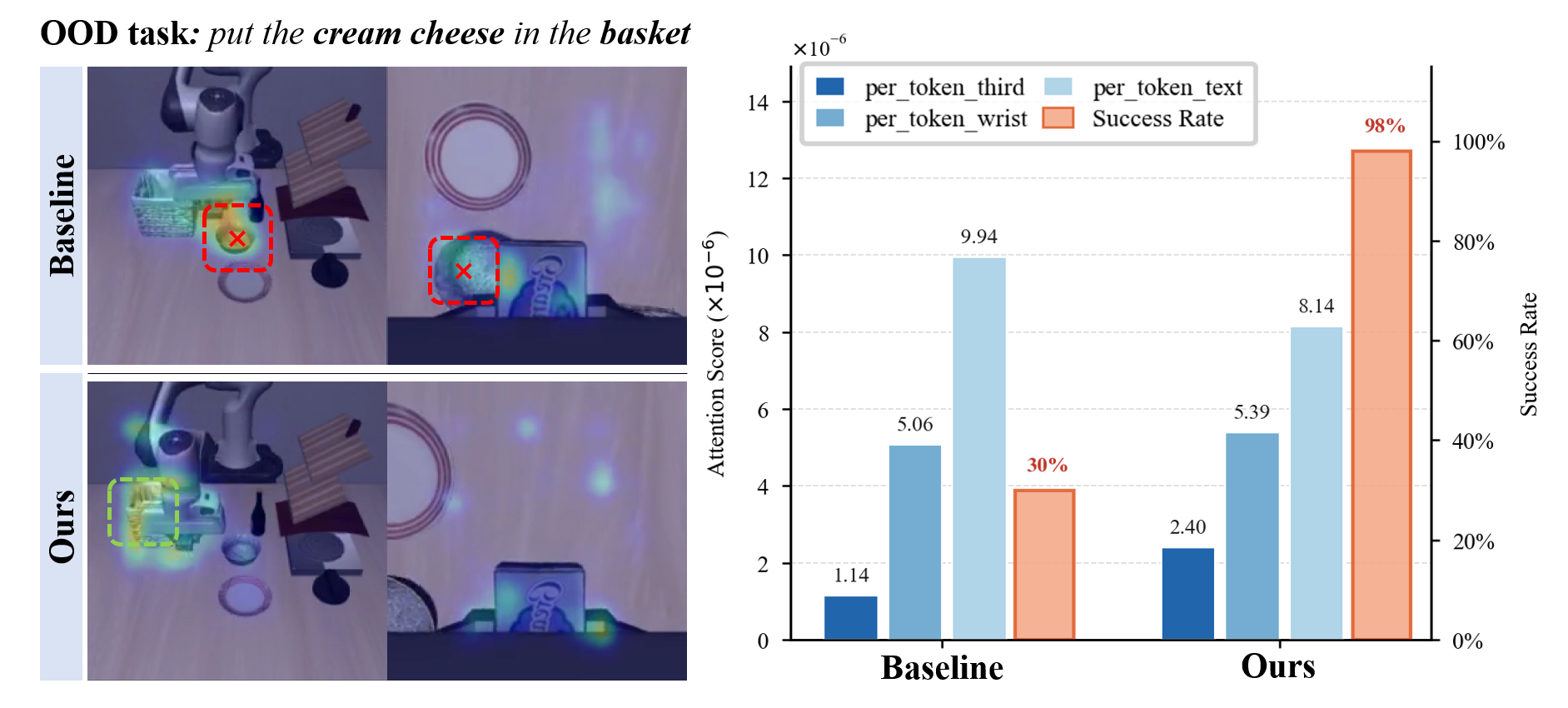}
    \caption{Comparison of attention maps between the vanilla $\pi_{0.5}$ and the model trained with asymmetric masking.}
    \label{fig:maskVis}
    \vspace{-0.5em}
\end{wrapfigure}

\textbf{Qualitative Analysis}. 
We further analyze the effect of AC-VLA by visualizing token-level attention scores on an OOD task, as shown in Figure~\ref{fig:maskVis}. In the attention maps of vanilla \(\pi_{0.5}\) (Baseline), the model exhibits a stark misalignment between attention and task intent, where attention is concentrated on an irrelevant target associated with in-distribution training. In contrast, the attention maps produced by our method are precise and task-aligned. The right panel of Figure~\ref{fig:maskVis} additionally reports the average token attention score during the ``place'' phase, produced by wrist view, third-person view, and text modality. We observe a larger than 100\% increase in per-token attention on the third-person view, confirming that AC-VLA effectively fosters global spatial grounding and compositional generalization.

\subsection{Evaluation on Real-World Scene}
    \label{sec:realworld}
\begin{figure}[t]
\centering
\begin{minipage}{0.6\linewidth}
\centering
\includegraphics[width=\linewidth]{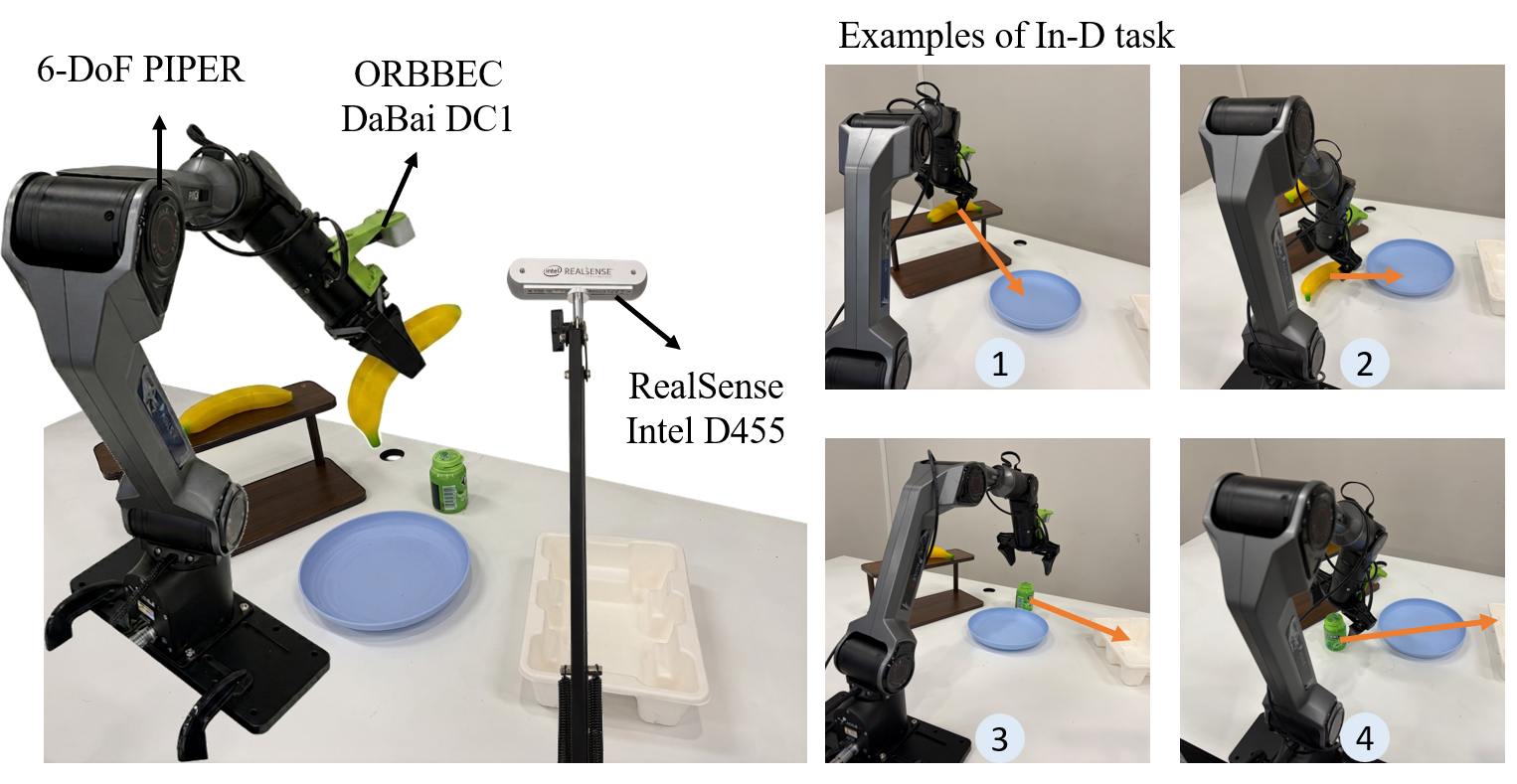} 
\caption{Real-world setup and raw task examples.}
\label{fig:real}
\end{minipage}
\begin{minipage}{0.38\linewidth}
\centering
    \captionof{table}{Real-world performance (\%) on In-D and OOD tasks.}
    \resizebox{\linewidth}{!}{
    \begin{tabular}{lc|c|c}
        \toprule
        \textbf{Model} & \textbf{In-D Tasks} & \textbf{OOD Tasks} &  \textbf{AVG} \\
        \midrule
       $\pi_{0.5}$                & \textbf{93.7}  & 35.0  &64.4\\
       \rowcolor{gray!20} \textbf{AC-VLA} & 88.7 & \textbf{82.5}\textcolor{red}{(+47.5)} & \textbf{85.6}\textcolor{red}{(+21.2)}\\
        \bottomrule
    \end{tabular}}
\label{tab:realTable}

\includegraphics[width=\linewidth]{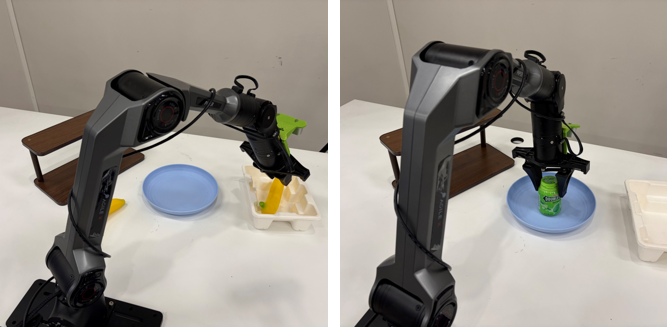} 
\captionof{figure}{Successful real-world OOD examples.}
\label{fig:real2}
\end{minipage}
\vspace{-1.5em}
\end{figure}

We further evaluate AC-VLA on a real-world robotic platform. As shown in Figure~\ref{fig:real}, the system consists of a 6-DoF PIPER arm, an ORBBEC DaBai DC1 wrist-view camera, and an Intel RealSense D455 third-person camera, forming a complete visual-manipulation pipeline. We design four in-distribution (In-D) tasks following the LIBERO-Goal setting: 
(1)\textit{pick up the banana between the wooden shelf and the plate and place it on the plate},
(2)\textit{pick up the banana on the wooden shelf and place it on the plate}, 
(3)\textit{pick up the chewing gum between the wooden shelf and the plate and place it in the pulp tray}, 
(4)\textit{pick up the chewing gum in front of the plate and place it in the pulp tray}. 
The two OOD tasks, illustrated in Figure~\ref{fig:real2}, are (1) \textit{put the banana on the wooden shelf in the pulp tray}, and (2) \textit{put the chewing gum in front of the plate and place it on the plate}.

The quantitative results are reported in Table 4. The baseline model \(\pi_{0.5}\) achieves 93.7\% success on In-D tasks but only 35\% on OOD tasks, resulting in an average performance of 64.35\%. In contrast, our AC-VLA maintains a high In-D success rate of 88.7\% while significantly boosting OOD performance to 82.5\%, yielding an overall average of 85.6\%. This demonstrates that our method retains strong in-distribution performance while substantially improving generalization to novel scenarios.

\section{Conclusion}
We propose AC-VLA, a plug-and-play Action Compositional learning framework that endows VLA models with robust compositional OOD generalization for robotic manipulation. By integrating two architecture-agnostic components--a compositional learning module that decomposes demonstrations into reusable sub-task units with mixed training, and a state-conditioned asymmetric masking strategy that suppresses wrist-view shortcuts--AC-VLA directly addresses the dual failure modes of trajectory overfitting and perceptual shortcuts. Our framework requires no architectural modification and can be seamlessly integrated into any VLA backbone. Extensive experiments on the LIBERO and LIBERO-OOD benchmarks demonstrate that AC-VLA achieves significant absolute improvement on compositional OOD tasks while maintaining near-perfect in-distribution performance. Real-world evaluations further validate the framework's practicality and effectiveness. 

\section{Limitations}
While AC-VLA achieves strong compositional generalization, it also has some limitations. First, the accuracy of offline task decomposition depends on both the LLM's semantic parsing capability and the fidelity of our proprioceptive-based aligner; for highly ambiguous or domain-specific instructions, the generated sub-task descriptions may be inaccurate, which could propagate errors into trajectory alignment and mixed training. Second, the benefits of explicit compositional learning may diminish when training on extremely large and diverse datasets that already cover most object–target combinations, effectively making the OOD gap negligible.


\clearpage
\acknowledgments{If a paper is accepted, the final camera-ready version will (and probably should) include acknowledgments. All acknowledgments go at the end of the paper, including thanks to reviewers who gave useful comments, to colleagues who contributed to the ideas, and to funding agencies and corporate sponsors that provided financial support.}


\bibliography{example}  

\end{document}